\documentclass{article}
\usepackage{spconf,amsmath,graphicx}

\usepackage{subfigure}
\usepackage{amssymb} 
\usepackage{amsthm}

\usepackage{hyperref}
\usepackage{color, colortbl, xcolor} 
\usepackage{diagbox} 
\usepackage{soul}

\newcommand{\sdkl}{\mathcal{D}_{\mathtt{KL}}}

\title{
\vspace{-2cm}{\hspace{-1cm}\footnotesize \textnormal{\textit{Paper accepted to IEEE ICASSP'2023, Greek island of Rhodes, 4-10 June 2023. \\{\bf LARGE FONTS}. The 22nd of February 2023.\\}}
}\vspace{+1.2cm}A study on the invariance in security whatever the dimension of images \\for the steganalysis by deep-learning}

\name{Kévin PLANOLLES$^{\star}$ \qquad Marc CHAUMONT$^{\star \dagger}$, Senior Member, IEEE \qquad Frédéric COMBY$^{\star}$ \thanks{The authors would like to thank the French Defense Procurement Agency (DGA) for its support through the ANR Alaska project (ANR-18-ASTR-0009). We also thank Douglas Benhamou and Mohamed Benkhettou for their technical help.}}
\address{$^{\star}$ LIRMM, Univ. Montpellier, CNRS 161 rue Ada, 34392 Montpellier Cedex 05, France \\ $^{\dagger}$ Univ. Nîmes Place Gabriel Péri, 30000 Nîmes Cedex 01, France}

\begin{document}
%
\maketitle
\begin{abstract}
In this paper, we study the performance invariance of convolutional neural networks when confronted with variable image sizes in the context of a more "wild steganalysis". First, we propose two algorithms and definitions for a fine experimental protocol with datasets owning "similar difficulty" and "similar security". The "smart crop 2" algorithm allows the introduction of the Nearly Nested Image Datasets (NNID) that ensure "a similar difficulty" between various datasets, and a dichotomous research algorithm allows a "similar security". Second, we show that invariance does not exist in state-of-the-art architectures. We also exhibit a difference in behavior depending on whether we test on images larger or smaller than the training images. Finally, based on the experiments, we propose to use the dilated convolution which leads to an improvement of a state-of-the-art architecture.
\end{abstract}
%
\section{Introduction}
\label{sec:intro}

Steganalysis is the subject of many academic publications but its use in real-world conditions ("in the wild") is often far from the laboratory protocols \cite{Ker2013_RealWorld} \cite{Cogranne2019_Alaska}. In laboratory conditions, the worst-case attack is used, meaning that: payload size, image development (demosaicing, gamma correction, blur, color balance, compression parameters, use of color, etc.), are known by the steganalyst.

This paper, therefore, fits in the spirit of works on a more "wild/realistic" steganalysis. We are specifically focusing on the case where the dimension\footnote{In the experiment we will build datasets all issued of {\em crop} from an initial dataset. Varying the dimension stands for a variation of width and height by cropping the images.} of the images is not known by the steganalyst. Said differently, we would like to keep {\em detection performances constant whatever the dimension} of the considered image. 

Since the advent of deep-learning in steganalysis in 2015 \cite{Chaumont2020}, a few architectures have the intrinsic capability to accept input images of variable size \cite{Yedroudj2018_Net}, \cite{Zhu-Net2019}, \cite{Reinel2021_GBRAS}, \cite{Fu2022_CCNet}, \cite{Su2021_EWNet}, \cite{Luo2022_Transfo}, \cite{FujiTsang2018_SID}, \cite{You2022_SiaSteg}, etc. Nevertheless, it is reasonable to ask ourselves if those architectures are efficient in detection whatever the dimension. More precisely we would like to know if the used architectures and/or the tricks ensure an {\em invariance in security}. This notion of {\em invariance in security} will be formally defined in the section \ref{ssec:tools}.

The contributions of that paper are: 1) a well thought experimental setting (dataset building, payload size tuning, and learning protocol) leading to the definition of {\em difficulty} and to the {\em Near-Nested Image Datasets (NNID)}, 2) the definition of the {\em invariance in security}, 3) the reported experimental observations and finally, 4) the proposition of an update of the architectures in order to be closer to invariance.  

In Section \ref{ssec:tools} we briefly discuss the payload size with a recall of the Square Root Law \cite{Ker2008_SRL}, and we present the state-of-the-art architectures able to process variable dimension images. In Section \ref{ssec:proposition}, we describe the way the NNID are built; this comes with the definitions of {\em "same" difficulty} of the datasets, and {\em "same" security} due to the embedding in each of those datasets. We also propose an upgrade to the deep-learning architectures. Then, in Section \ref{ssec:results} we give the experimental protocol, and the experimental results, and discussed the {\em invariance} notion.

\section{Tools used for our study}
\label{ssec:tools}

\subsection{Few words on the square root law}

It is well known that the size of a cover object is a major factor in its capacity to hide information, but the relationship between {\it secure steganographic capacity} and cover size is still a discussed subject. In Ker {\it et al.}'s paper  \cite{Ker2008_SRL}, it is shown that payload size should be proportional to the \emph{square root} of the cover size. From the Square Root Law we can derive, given a positive constant $k \in \mathbb{R}^+$, that the relative payload size, $\alpha$, for an adaptive embedding, ensuring the {\em same security} whatever the dimension $w\times h$ of an image, should be (in bit per pixel):
\begin{equation}\label{eq:SquareRootLaw}
\alpha = \frac{k}{wh} \times \sqrt{wh}\times \log(wh).
\end{equation}

But in practice, the statistical properties of the cover has to be taken into account, and it appears that the square root law is not usable as it is. For example, a homogeneous cover will be much less secure than a cover with a lot of textures. 

In our study, we propose another solution to choose the relative payload ensuring a {\em similar security} considering the dimension of images. It is based on careful datasets building and the use of a detector i.e. a deep-learning network (see Section \ref{ssec:proposition}). 


\subsection{State-of-the-art deep learning architectures}

There are roughly two families of deep learning architecture that have the property to accept images of various dimensions, those only using the average, and those using more than only one statistical moment.

For the family based on the average, we can mention the following networks. The Yedroudj-Net \cite{Yedroudj2018_Net} architecture is based on the Global Average Pooling (GAP) technique to aggregate the values of the feature maps just before feeding the classification block. 
The Zhu-Net \cite{Zhu-Net2019} architecture is inspired by Yedroudj-Net. Its main characteristic is the use of a Pyramidal Global Average pooling to retrieve features at multiple scales. A few other "small" architectures such as GBRAS-Net \cite{Reinel2021_GBRAS}, CC-Net \cite{Fu2022_CCNet}, etc, dealing with spatial image steganalysis have then improved the Yedroudj-Net and the Zhu-Net, and are also integrating a GAP. We should also mention a very recent proposition integrating a transformer \cite{Luo2022_Transfo} with a GAP. Finally, EWNet \cite{Su2021_EWNet} is an original approach that uses a coder-decoder, and proposes to compute the average of the image of scores for the cover (resp. the stego). 

For the family based on more than only one statistical moment, we can mention the following networks. The SID \cite{FujiTsang2018_SID} architecture is based on a modified version of the Ye-Net network \cite{Ye2017}. The main idea of the SID architecture is to extract four statistical moments (minimum, maximum, average, and variance) from the last feature maps and use those moments for classification. The SiaSteg \cite{You2022_SiaSteg} architecture is also based on the extraction of statistical moments (the same as SID), and uses a siamese network in order to add a contrastive loss in conjunction with the classification loss.  

In some of those papers, there are experiments on the "robustness" to the variation in dimension, and for a subset of those papers, the notion of {\it secure steganographic capacity} is mentioned, but in none of them the results agreed with this secure steganographic capacity notion. Additionally, none of them has set a {\em common difficulty} (see definition in section \ref{ssec:NNID}) between the dimensions, which ensures that the variation between datasets is only due to the relative payload size. In this paper, our experimental protocol ensures the {\em same empirical security} (thanks to adapted payloads) between the dimensions and works with datasets of various dimensions and {\em similar difficulty}.

\section{Proposition}
\label{ssec:proposition}

\subsection{Nearly-Nested Image Datasets (NNID) ensuring the {\bf {\em same difficulty}}} \label{gigognes}
\label{ssec:NNID}

For mastered experiments, we need to build what we name  {\em Nearly-Nested Image Datasets} (NNID) such that each dataset owns images of the same dimension, and each dataset is issued from a {\it cropped version} of the images belonging to the dataset with the biggest dimensions. This last dataset is named {\em mother dataset} and the images are named {\it mother} images.  

By using NNID we ensure that the development is the same in all of the datasets. We also impose, as an additional constraint, that the {\em difficulty} of each dataset is the same. By {\em same difficulty} we mean that the distribution of costs is the same whatever the dataset. This additional constraint implies a specific way to crop the images, and most importantly, ensure that the experimental results obtained between the various dimension will be comparable since the source cost distribution of each dataset is the same. With the NNID we are able to avoid any impact of the development or the difficulty, on the experimental results; All the datasets are very similar except for the dimension.

We use, as a {\em mother dataset}, the LSSD dataset \cite{Ruiz2021_LSSD}. LSSD is a mix of RAW images from ALASKA\#2, BOSS, StegoApp DB, Wesaturate, RAISE, and Dresden datasets and uses a modified development script issued from the Alaska competition\footnote{See \url{https://www.lirmm.fr/~chaumont/LSSD.html}.}.

For a given image from the {\em mother dataset}, we define the {\em smart crop 2} as the crop (i.e the area of the {\it mother image}) that keeps the same distribution of costs between the {\em mother} image and the cropped one. 
To compare the distribution, we used symmetrized Kullback-Leibler distance: 
\begin{equation}
\sdkl(P,Q) := \sum_i P(i) \log \frac{P(i)}{Q(i)} + \sum_i Q(i) \log \frac{Q(i)}{P(i)},
\end{equation}
for given discrete probability distributions $P$ and $Q$. Note that in this paper, we consider the cost obtained with the S-UNIWARD algorithm \cite{Holub2014_UNIWARD}, which is one of the most efficient and generic embedding algorithm \cite{Sedighi2016_TossBOSS}. 


Note that a brute force approach has to be done to test all the areas and find the one with the minimal Kullback-Leibler symmetrized distance\footnote{For the NNID, the search space is square areas of dimensions 256x256, 512x512, 1024x1024, and 2048x2048.}. To avoid redundant computation and reduce the computational cost, we use the {\em histogram integral} approach \cite{Porikli2005_integral_histogram} which is an extension of the well-known {\em integral image} approach, defined in the paper of Viola and Jones \cite{Viola2001_integral_image}. Given a mother image of size $n\times m$, and a crop area of size $w\times w$, the complexity for computing all the histograms for all the positions in the mother image is O($n \times m \times w^2$) whereas it is a much smaller complexity of O($n\times m$) with the histogram integral approach. 

\begin{figure}[htb!]
\includegraphics[width=0.11\textwidth]{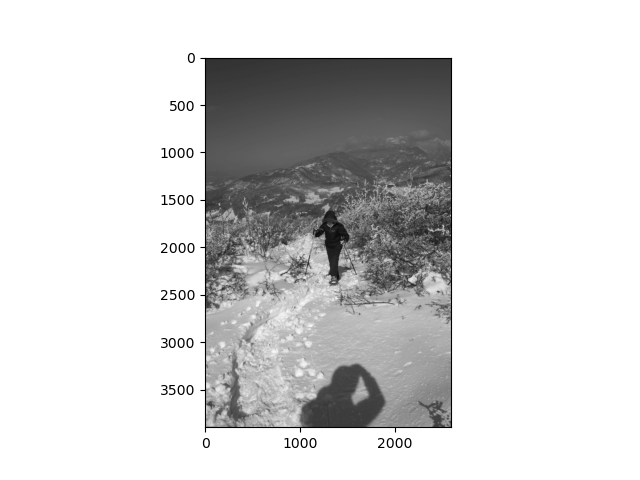}
\includegraphics[width=0.11\textwidth]{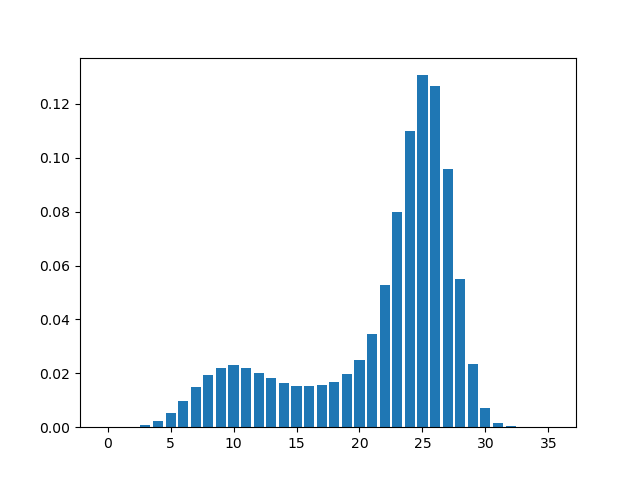}
\includegraphics[width=0.11\textwidth]{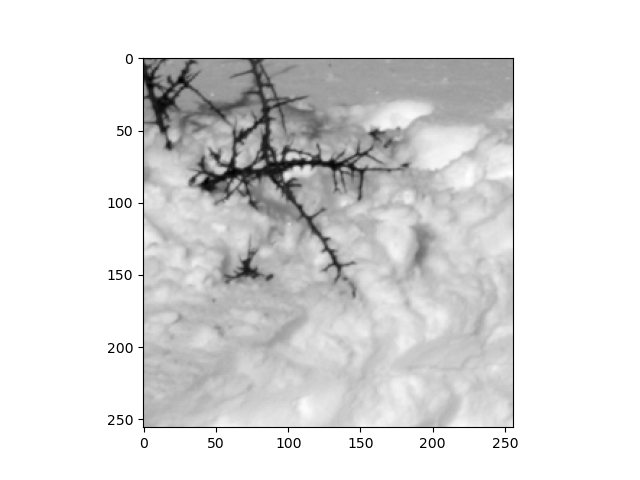}
\includegraphics[width=0.11\textwidth]{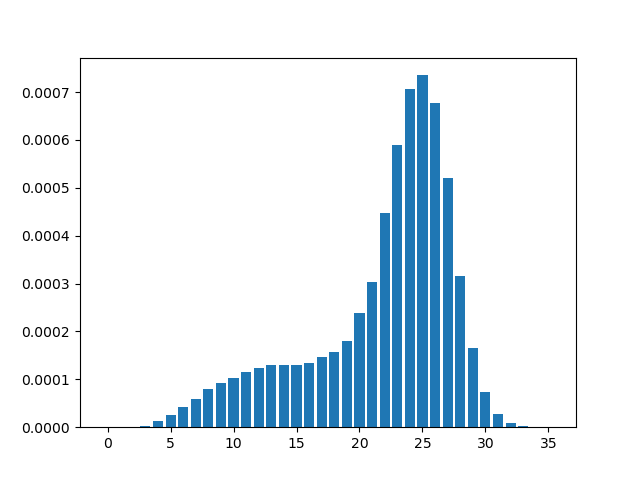}
\caption{A {\em mother image} of size $2594\times3898$ with its costs histogram, and a near-nested image of size $256\times 256$, issued from the {\em smart crop 2} of the {\em mother image}, with its costs histogram.}
\label{fig:costHist}
\end{figure}

Figure \ref{fig:costHist} shows a {\em mother image} of size $2594\times3898$ with its costs histogram, and a near-nested image of size $256\times 256$, issued from the {\em smart crop 2} of the {\em mother image}, with its costs histogram.


\subsection{Relative payload in each dataset ensuring the {\bf {\em same security}}}
\label{ssec:payload}

Given the NNID and an embedding algorithm, we need to find the number of bits to embed in each dataset. In this paper, we only need to focus on the average accuracy obtained by a classifier for each dataset. Using the square root law in order to find the correct payload size to embed in each dataset (such that the security is the same for all) does not ensure, in practice, the same accuracy for all the datasets.  

In order to obtain the relative payload size to embed for each dimension, we thus go by a dichotomous method, by running, for each dimension, multiple detections until finding the desired accuracy. Note that we use the Square Root Law as an initialization of this empirical research, for finding the relative payload size for each dimension (see Equation \ref{eq:SquareRootLaw}). 
This protocol is the one that was used to set the payload sizes in the experiments of this article.

\subsection{Invariance in security}

We give here the definition of the {\em invariance in security}. Let's suppose an NNID that ensures similar pixel distributions, similar contents, the same development, and the {\em same difficulty} for all the datasets, plus a payload size for the embedding (described in Section \ref{ssec:payload}) that ensures the same {\em empirical security}. We define a deep learning network \emph{invariant in security with respect to the dimension} when its obtained average accuracy is the same whatever the dimensions. This definition, even if very simple, allows a much finer analysis of the invariance property than done previously. 




\subsection{Use of dilated convolution}
\label{ssec:dilated}

Some authors have proposed to introduce invariance to the dimension by performing multiple convolutions in parallel on resized versions of the input image \cite{Noord2017_Invariance} \cite{Marcos2018_equivariance}. Nevertheless, we cannot perform resampling without losing information on the steganographic noise. We, therefore, propose to use the principle of dilated convolution \cite{Yu2015_Dilated}, which consists in spacing the elements of the convolution kernel, as an approximation of the resizing. There is then no more subsampling of the input image.

The dilated convolution operator is defined, for an image, ${\bf z}: \mathbb{Z}^2 \mapsto \mathbb{R}$, a kernel, ${\bf k}: \mathbb{Z}^2 \mapsto \mathbb{R}$, and a scalar, $d \in \mathbb{N}$, standing for the dilation factor, by \cite{Yu2015_Dilated}:
\begin{equation}
({\bf z}*{\bf k})(x,y) = \sum_i \sum_j {\bf z}(x-d\cdot i, y-d\cdot j){\bf k}(i,j).
\nonumber
\end{equation}

Lots of the networks designed for steganalysis have their first convolution block that serves to remove the influence of the image content \cite{Chaumont2020}. Additionally, most of them have their second convolution block that outputs feature maps whose height and width are equal to these of the image fed to the network. 

One thus can modify most of the networks by substituting the second convolution block with an inception block \cite{Szegedy2015_GoogleNet} made of dilated convolutions of various dilation factors. This can be easily done without changing neither the number of parameters of the network, either the depth, width, or height of the feature maps.

\begin{figure}[htb!]
\centering
\includegraphics[width=0.45\textwidth]{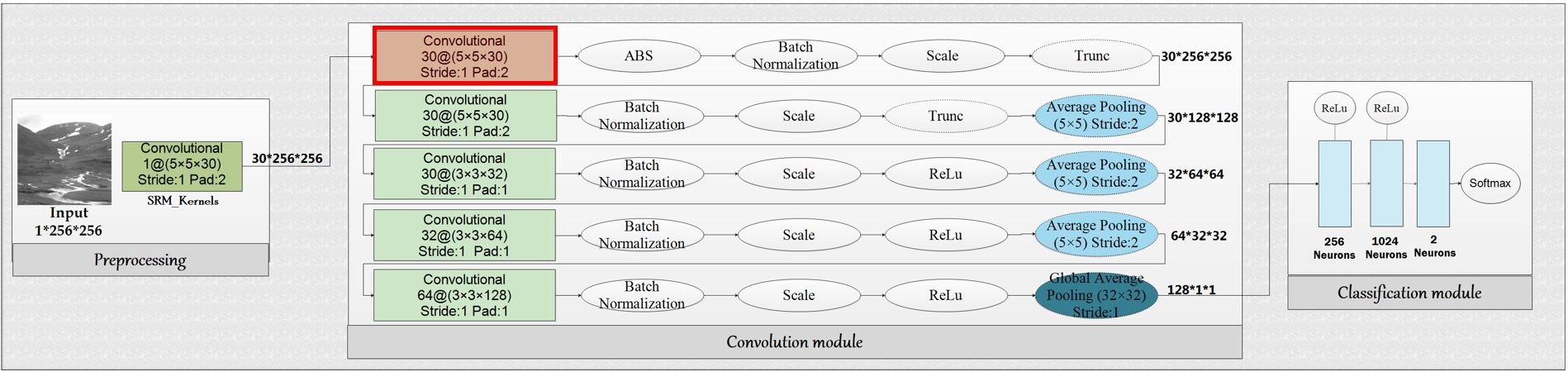}
\caption{The Yedroudj-Net architecture with the block where dilated convolutions are applied is highlighted.} 
\label{fig:yedroudj}
\end{figure}

Applied to the Yedroudj-Net architecture (see Figure \ref{fig:yedroudj}), the second convolution block still inputs and outputs 30 features maps of size $256\times256$, but the 30 convolution filters of size $5\times5$ are replaced by 10 standard convolutions (without dilatation), 10 convolutions with dilation of 2 (i.e. the kernel elements are spaced by one pixel), and 10 convolutions with dilation 4 (i.e. the elements of the kernel are spaced by three pixels), all of them of size $5\times5$.   

\section{Experiments}
\label{ssec:results}

\subsection{Experimental protocol}

In the NNID\footnote{Send an email to \href{mailto:marc.chaumont@lirmm.fr}{marc.chaumont@lirmm.fr} to get a private copy.}, each dataset contains images whose dimensions are all equal. Each of them is named UNI\_{$w$}, with  $w$ $\in \{256, 512, 1024,$ or $ 2048\}$, the size $w\times w$ of the images of the dataset. UNI stands for unidimensional.

For each UNI dataset, the train part is made of 12000 cover/stego pairs (24000 images in total) with 19200 images dedicated to the train, and 4800 images to the validation. For each UNI dataset, the test dataset is made of 3000 cover/stego pairs (6000 images in total). 

An additional dataset, named MULTI, is derived from the datasets UNI\_$w$, with $w$ $\in \{256, 512, 1024\}$ such that 
4000 cover/ste\-go pairs of each UNI dataset are randomly selected, in order to obtain 12000 cover/stego pairs, and thus a total of 24000 images, with 19200 images dedicated to the train, and 4800 images to the validation. 

We used S-UNIWARD\cite{Holub2014_UNIWARD} for the embedding in the spatial domain with its MATLAB implementation\footnote{See \url{http://dde.binghamton.edu/download/stego_algorithms/}.}. 

We trained the SID network\footnote{SID is from the Github of SiaSteg: \url{https://github.com/SiaStg/SiaStegNet}} and the Yedroudj-Net network\footnote{Yedrouj-Net: \href{https://github.com/yedmed/steganalysis_with_CNN_Yedroudj-Net}{https://github.com/yedmed/steganalysis\_with...}}
using the hyper-parameters from the GitHubs, with a batch of 32. We trained the "Dilated" Yedroudj-Net\footnote{Dilated Yedroudj-Net will be soon on a GitHub.} with the same parameters as Yedroudj-Net, except that the batch size is set to 16.



For all networks, the learning rate is divided by two if the validation's accuracy does not improve for ten consecutive epochs. We set the maximum number of epochs to 400 for all networks, but if the validation's accuracy does not improve for fifty consecutive epochs, the learning is stopped. After the training, we saved the weights which gave the best validation accuracy and use them for the test on the test dataset (The standard deviation of networks is extremely low; As an example, for a SID trained on $256\times256$ images it is below 0.02\% when testing the top-5 trained networks). 

Probably due to unadapted hyper-parameters, SiaSteg \footnote{SiaSteg: \url{https://github.com/SiaStg/SiaStegNet}}, and Zhu-Net \footnote{Zhu-Net: \url{https://github.com/1204BUPT/Zhu-Net-image-steganalysis}} networks did not converge. We thus restricted our study to SID and Yedroudj-Net which are two representatives of each family of approaches.

\subsection{Relative payload for the UNI datasets}

As a preliminary experiment, we embedded at a relative payload size of 0.4 bpp (bits per pixels) in the $256 \times 256$ images, and deduced with the Square Root Law mentioned previously, a relative payload of 0.225 (resp. 0.125, and 0.06875) for $512\times 512$ (resp. $1024\times 1024$, and $2048\times 2048$) images. We then trained and tested a SID on $256\times256$ images, and another one on $512\times512$ images. Accuracies were not at all equal with respectively 69\% and 62\%, indicating that some hypotheses (asymptotic hypothesis + independencies assumption) of the Square Root Law were not met and that this law could not be used given our experimental conditions.

We thus updated the relative payload size used for each dimension as explained in section \ref{ssec:payload} such that after a manual dichotomous approach the accuracy for each UNI dataset for the Yedroudj-Net model is 76\%. Note that this accuracy is a meaningful accuracy that lets a sufficient margin for future experiments.

The table \ref{payload_adj} gives the adjusted payload for each dimension and the corresponding accuracy.
\begin{table}
\centering
\caption{Relative payload for each dimension.
\label{payload_adj}}
\begin{tabular}{|c|c|c|}
\hline 
Dimension & Relative payload & Accuracy (Yedroudj-Net) \\ 
\hline 
256 & 0.4 & 76.97\% \\ 
\hline 
512 & 0.3204 & 76.38\% \\ 
\hline 
1024 & 0.28895 & 76.78\% \\ 
\hline
\end{tabular}
\end{table}

\subsection{Learn on a UNI dataset}

Our first case study consists to learn a network on a UNI dataset (i.e. images have the same dimensions) and evaluating its invariance in security on the other UNI dataset. We note as $\mathtt{ARCHI}$-$\mathtt{DIM}$, where $\mathtt{ARCHI}$ denotes the name of the architecture, $\mathtt{DIM}$ denotes the dimension on which the network was trained.  

The results for SID, Yedroudj-Net, noted \texttt{Y}, and Dilated-Yedroudj-Net, noted \texttt{DY}, are reported in Table \ref{tab::uni}. The diagonals correspond to the clairvoyant scenario without any mismatch in dimension. The accuracy for the diagonal values for SID (resp. DY) are close. It confirms that the security by accuracy defined in Section \ref{ssec:payload} is a fine grain definition of the security since given a detector, for each UNI dataset it gave close accuracy (close to 69.9\% for SID and close to 77.5\% for DY).

\begin{table}[htb!]
\centering
\caption{Accuracies for the SID, the Yedroudj-Net (noted \texttt{Y}), and the Dilated-Yedroudj-Net (noted \texttt{DY}), models. \label{tab::uni}}
\begin{tabular}{|c|c|c|c|}
\hline 
\textbf{Dim} & \textbf{\texttt{SID-256}} & \textbf{\texttt{SID-512}} & \textbf{\texttt{SID-1024}} \\
\hline 
$256 \times 256$ & \cellcolor[gray]{.95}\bf 69.48\% & 67.05\% & 60,9\% \\ 
\hline 
$512 \times 512$ & 69.30\% & \cellcolor[gray]{.95}\bf 70.7\% & 66.93\% \\ 
\hline 
$1024 \times 1024$ & 66.73\% & 66.93\% & \cellcolor[gray]{.95} \bf 69.62\% \\ 
\hline 
\hline
\textbf{Dim} & \textbf{\texttt{Y-256}} & \textbf{\texttt{Y-512}} & \textbf{\texttt{Y-1024}} \\
\hline
$256 \times 256$ & \cellcolor[gray]{.95} \bf 76.97\% & 73.48\% & 71.76\% \\
\hline
$512 \times 512$ & 74.55\% & \cellcolor[gray]{.95} \bf 76.38\% & 74.97\% \\
\hline
$1024 \times 1024$ & 72.83\% & 73.57\% & \cellcolor[gray]{.95} \bf 76.78\% \\
\hline
\hline 
\textbf{Dim} & \textbf{\texttt{DY-256}} & \textbf{\texttt{DY-512}} & \textbf{\texttt{DY-1024}}\\ 
\hline 
$256 \times 256$ & \cellcolor[gray]{.95} \bf 77.7\% & 76.25\% & 71.92\% \\ 
\hline 
$512 \times 512$ & 75.21\% & \cellcolor[gray]{.95} \bf 77.3\% & 76.2\% \\ 
\hline 
$1024 \times 1024$ & 72.03\% & 76.88\% & \cellcolor[gray]{.95} \bf 77.53\% \\ 
\hline
\end{tabular}
\label{tab:UNI}
\end{table}

Looking at the non-diagonal results it appears that the performance systematically decrease compared to the diagonal. The biggest performance loss is for the learning at 1024 when evaluated on smaller dimensions, with a loss from 8\% (SID) to 5\% (Yedroudj-Net and Dilated-Yedroudj-Net).  

SID and Yedroudj-Net are two representatives of the two families of networks accepting images of various dimensions and are also well-established networks. The above results let us guess that current networks from the literature are not intrinsically invariant to the change in dimension since their accuracies, when confronted with dimension never seen during the learning, are not constant.  

Going deeper into the observation of the variance in security phenomenon, considering for example the DY-512, (see table \ref{tab:UNI}), it appears that the accuracies are very similar (around 76\%) for a test on the lower dimension $256\times256$ or on the higher dimension $1024\times1024$, but the behavior varies greatly with image size. Indeed, as observed in Table \ref{mat_conf_dyn512}, when tested on a lower dimension, the model tends to over-classify in stego, while when tested on a higher dimension, it over-classifies images as covers.

\begin{table}[htb!]
\centering
\caption{Confusion matrices of \texttt{DY-512}.\label{mat_conf_dyn512}}
\begin{tabular}{|c|c|c||c|c|c|}
\multicolumn{3}{l}{\small  DY-512 tested on $256 \times 256$} & \multicolumn{3}{l}{\small  DY-512 tested on $1024 \times 1024$}\\
\hline
\diagbox{\scriptsize Truth}{\scriptsize Pred.} & \scriptsize Cover &  \scriptsize Stego & \diagbox{\scriptsize Truth}{\scriptsize Pred.} & \scriptsize Cover & \scriptsize Stego \\ 
\hline 
\scriptsize Cover & \scriptsize 2140 & \scriptsize 860 &	\scriptsize Cover & \scriptsize 2530 & \scriptsize 470 \\  
\hline 
\scriptsize Stego & \scriptsize 565 & \scriptsize 2435 & \scriptsize Stego & \scriptsize 917 & \scriptsize 2083 \\  
\hline
\scriptsize Percentage & \scriptsize 45\% & \scriptsize 55\% & \scriptsize Percentage &  \scriptsize 56\% &  \scriptsize 43\%\\
\hline
\end{tabular}
\end{table}

We checked if this variance in security was due to a wrong decision threshold that should be updated when facing a dimension not seen during the learning. It appears in experiments with Y-512 and SID-512 that updating the threshold does not change the accuracy for the unseen dimensions. 

We also checked if there was a difference in the noise pattern depending on the dimension which could explain the variance in performance. Using the activation map (using integrated gradient \cite{Sundararajan2017_IR}) for a few stego images of size $512\times 512$, and the associated nearly nested images $256\times256$, such that the first group is well classified and the second group is wrongly classified, it appears that the activation maps are very close. Thus the latent space may be very sensitive to small differences indicating that the feature maps are not sufficiently robust (i.e varies) to the change in dimension.


Finally, we should mention that the introduction of the dilated convolution to the Yedroudj-Net (see Table \ref{tab::uni}) increases the accuracy by 1\% (see diagonal results) and slightly improves the results on unseen dimensions (non-diagonal results), so probably encouraging the invariance to the dimension.

\subsection{Learn on the MULTI dataset}

Our second case study consists of learning a network on the MULTI dataset (i.e. images have various dimensions) and evaluating its invariance in performance on the UNI datasets. We refer as $\mathtt{ARCHI}$-\texttt{MULTI}, the networks trained on multiple dimensions,  where $\mathtt{ARCHI}$ denotes the name of the architecture.

The results for SID, Yedroudj-Net, and Dilated-Yedroudj-Net are reported in Table \ref{tab:multi}. While those models are not invariant with respect to dimension, the variations in accuracies are less important compared to learning on a UNI dataset, particularly for $512 \times 512$ and $1024 \times 1024$ images. This confirms that training on multiple dimensions improves the invariance, which was a trick already used in past publications, as a measure for improving the robustness to various dimensions. 

Nevertheless, as we can see in Table \ref{tab:multi}, the invariance is not reached since the accuracy for the dimension $256\times256$ is lower by 1.6\% to 3.7\% than those obtained at dimension $512\times512$ or $1024\times1024$. This phenomenon is also observed with the Dilated-Yedroudj-Net-MULTI which highlights the fact that future work has to be done to ensure invariance. We should also note that the Dilated-Yedroudj-Net-MULTI allows obtaining the best results which is a piece of interesting practical information for the topic of invariance in dimension.  

\begin{table}[htb!]
\caption{\texttt{SID-MULTI}, \texttt{Y-MULTI} and \texttt{DY-MULTI} accuracies.\label{tab:multi}}
\begin{tabular}{|c|c|c|c|}
\hline 
\textbf{Dim} & \bf \texttt{SID-MULTI}& \bf \texttt{Y-MULTI} & \bf \texttt{DY-MULTI}\\ 
\hline 
$256 \times 256$ & 66.93\% & 73.93\% & 75.63\% \\ 
\hline 
$512 \times 512$ & 69.46\% & 75.5\% & 78.1\% \\ 
\hline 
$1024 \times 1024$ & 70.6\% & 75\% & 78.06\% \\ 
\hline 
\end{tabular}
\end{table}

\vspace{-0.4cm}
\section{Conclusions}\label{sect:chap5_conclu}
\vspace{-0.1cm}

In this paper, we explain how to build the Near Nested Image Datasets (NNID) thanks to a {\it smart crop 2} applied to an initial dataset. Those NNI Datasets, all each containing images of the same dimension, have similar pixel distribution, similar semantic content, same development, and the same distribution of costs such that we talk about {\em same difficulty} for all of the datasets. We also propose to dichotomously find a relative payload size for each dataset such that a pre-chosen neural network exhibit the same accuracies, leading to the notion of {\em same security} for all the cover/stego datasets. During the building of the NNI Datasets, we thus exhibit that the theoretical results of the square root law might not apply in a real-world context.  

Then, using the resulting cover/stego datasets allows us to observe that even in the case of learning with images of various dimensions, there is not any property of {\em invariance in security} for the current state-of-the approaches. We also remarked a difference in behavior depending on whether the network was tested on a larger or smaller dimension than the one on which it was trained. In the case of a lower dimension, there is an overestimation of the number of stegos while in the case of a higher dimension, there is an overestimation of the number of covers. Finally, we proposed a new architecture, Dilated-Yedroudj-Net, which gave better results than the other networks.\footnote{FAQ can be found there \url{https://www.lirmm.fr/~chaumont/publications/QA-ICASSP2023.pdf}}


\bibliographystyle{IEEEbib}
\bibliography{bib}

\begin{thebibliography}{10}

\bibitem{Ker2013_RealWorld}
Andrew.~D. Ker, Patrick Bas, Rainer B\"{o}hme, R{\'e}mi Cogranne, Scott Craver,
  Tomas Filler, Jessica Fridrich, and Tomas Pevn\'{y},
\newblock ``{Moving Steganography and Steganalysis from the Laboratory into the
  Real World},''
\newblock in {\em Proceedings of the 1st ACM Workshop on Information Hiding and
  Multimedia Security, IH\&MMSec'2013}, Montpellier, France, June 2013, pp.
  45--58.

\bibitem{Cogranne2019_Alaska}
R{\'e}mi Cogranne, Quentin Giboulot, and Patrick Bas,
\newblock ``{The ALASKA Steganalysis Challenge: A First Step Towards
  Steganalysis},''
\newblock in {\em Proceedings of the ACM Workshop on Information Hiding and
  Multimedia Security}, Paris, France, July 2019, IH\&MMSec'2019, pp. 125--137.

\bibitem{Chaumont2020}
Marc Chaumont,
\newblock ``{Deep Learning in steganography and steganalysis},''
\newblock in {\em {Digital Media Steganography: Principles, Algorithms,
  Advances}}, M.~Hassaballah, Ed., chapter~14, pp. 321--349. Elsevier, July
  2020.

\bibitem{Yedroudj2018_Net}
Mehdi Yedroudj, Fr\'ed\'eric Comby, and Marc Chaumont,
\newblock ``{Yedrouj-Net: An Efficient CNN for Spatial Steganalysis},''
\newblock in {\em Proceedings of the IEEE International Conference on
  Acoustics, Speech and Signal Processing, ICASSP'2018}, Calgary, Alberta,
  Canada, Apr. 2018, pp. 2092--2096.

\bibitem{Zhu-Net2019}
Ru~Zhang, Feng Zhu, Jianyi Liu, and Gongshen Liu,
\newblock ``{Depth-Wise Separable Convolutions and Multi-Level Pooling for an
  Efficient Spatial CNN-Based Steganalysis; ({\it previously named "efficient
  feature learning and multi-size image steganalysis based on cnn" on
  ArXiv})},''
\newblock {\em IEEE Transactions on Information Forensics and Security, TIFS},
  vol. 15, pp. 1138--1150, 2020.

\bibitem{Reinel2021_GBRAS}
Tabares-Soto Reinel, Arteaga-Arteaga~Harold Brayan, Bravo-Ortiz~Mario
  Alejandro, Mora-Rubio Alejandro, Arias-Garzón Daniel, Alzate-Grisales~Jesús
  Alejandro, Burbano-Jacome~Alejandro Buenaventura, Orozco-Arias Simon, Isaza
  Gustavo, and Ramos-Pollán Raúl,
\newblock ``{GBRAS-Net: A Convolutional Neural Network Architecture for Spatial
  Image Steganalysis},''
\newblock {\em IEEE Access}, vol. 9, pp. 14340--14350, 2021.

\bibitem{Fu2022_CCNet}
Tong Fu, Liquan Chen, Zhangjie Fu, Kunliang Yu, and Yu~Wang,
\newblock ``{CCNet: CNN Model with Channel Attention and Convolutional Pooling
  Mechanism for Spatial Image Steganalysis},''
\newblock {\em Journal of Visual Communication and Image Representation}, vol.
  88, Oct. 2022.

\bibitem{Su2021_EWNet}
Ante Su, Xianfeng Zhao, and Xiaolei He,
\newblock ``{Arbitrary-Sized JPEG Steganalysis Based on Fully Convolutional
  Network},''
\newblock in {\em Proceedings of the International Workshop on
  Digital-forensics and Watermarking, IWDW'2021}, Beijing, China, Nov. 2021, p.
  197–211, Springer-Verlag.

\bibitem{Luo2022_Transfo}
Ge~Luo, Ping Wei, Shuwen Zhu, Xinpeng Zhang, Zhenxing Qian, and Sheng Li,
\newblock ``{Image Steganalysis with Convolutional Vision Transformer},''
\newblock in {\em Proceedings of the IEEE International Conference on
  Acoustics, Speech and Signal Processing, ICASSP'2022}, Marina Bay Sands,
  Singapore, May 2022, pp. 3089--3093.

\bibitem{FujiTsang2018_SID}
Cl{\'e}ment~Fuji Tsang and Jessica~J. Fridrich,
\newblock ``{Steganalyzing Images of Arbitrary Size with CNNs},''
\newblock in {\em Proceedings of the IS\&T, Electronic Imaging, Media
  Watermarking, Security, and Forensics, MWSF'2018}, San Francisco, CA, Feb.
  2018, vol. 2018, pp. 121--1--121--8.

\bibitem{You2022_SiaSteg}
Weike You, Hong Zhang, and Xianfeng Zhao,
\newblock ``{A Siamese CNN for Image Steganalysis},''
\newblock {\em IEEE Transactions on Information Forensics and Security}, vol.
  16, pp. 291--306, 2021.

\bibitem{Ker2008_SRL}
Andrew~D. Ker, Tom\'{a}\v{s} Pevn\'{y}, Jan Kodovsk\'{y}, and Jessica Fridrich,
\newblock ``{The Square Root Law of Steganographic Capacity},''
\newblock in {\em Proceedings of the 10th ACM Workshop on Multimedia and
  Security}, New York, NY, USA, 2008, MM\&Sec'2008, p. 107–116, Association
  for Computing Machinery.

\bibitem{Ye2017}
Jian Ye, Jiangqun Ni, and Y.~Yi,
\newblock ``{Deep Learning Hierarchical Representations for Image
  Steganalysis},''
\newblock {\em IEEE Transactions on Information Forensics and Security, TIFS},
  vol. 12, no. 11, pp. 2545--2557, Nov. 2017.

\bibitem{Ruiz2021_LSSD}
Hugo Ruiz, Mehdi Yedroudj, Marc Chaumont, Fr{\'e}d{\'e}ric Comby, and
  G{\'e}rard Subsol,
\newblock ``{LSSD: a Controlled Large JPEG Image Database for
  Deep-Learning-based Steganalysis ''into the Wild''},''
\newblock in {\em Proceedings of the 25th International Conference on Pattern
  Recognition, ICPR'2020, Workshop on MultiMedia FORensics in the WILD,
  MMForWILD'2020}, Virtual (formerly Milan), Italy, Jan. 2021, vol. 12666 of
  {\em Lecture Notes in Computer Science}, pp. 470--483, Springer International
  Publishing.

\bibitem{Holub2014_UNIWARD}
Vojtech Holub, Jessica Fridrich, and Tomas Denemark,
\newblock ``{Universal Distortion Function for Steganography in an Arbitrary
  Domain},''
\newblock {\em EURASIP Journal on Information Security, JIS}, vol. 2014, no. 1,
  2014.

\bibitem{Sedighi2016_TossBOSS}
Vahid Sedighi, Jessica~J. Fridrich, and R{\'{e}}mi Cogranne,
\newblock ``{Toss that BOSSbase, Alice!},''
\newblock in {\em Proceedings of the Media Watermarking, Security, and
  Forensics, MWSF'2018, Part of IS\&T International Symposium on Electronic
  Imaging, EI'2016}, San Francisco, California, USA, Feb. 2016, pp. 1--9.

\bibitem{Porikli2005_integral_histogram}
Fatih~Murat Porikli,
\newblock ``{Integral Histogram: A Fast Way To Extract Histograms in Cartesian
  Spaces},''
\newblock in {\em {Proceedings of the IEEE Computer Society Conference on
  Computer Vision and Pattern Recognition, CVPR'2005}}, San Diego, CA, USA,
  June 2005, vol.~1, pp. 829--836.

\bibitem{Viola2001_integral_image}
Paul~A. Viola and Michael~J. Jones,
\newblock ``{Rapid Object Detection using a Boosted Cascade of Simple
  Features},''
\newblock in {\em Proceedings of the IEEE Computer Society Conference on
  Computer Vision and Pattern Recognition, CVPR'2001}, Kauai,HI, {USA}, Dec.
  2001, vol.~1, pp. 511--518.

\bibitem{Noord2017_Invariance}
Nanne van Noord and Eric~O. Postma,
\newblock ``{Learning Scale-Variant and Scale-Invariant Features for Deep Image
  Classification},''
\newblock {\em Pattern Recognition}, vol. 61, pp. 583--592, 2017.

\bibitem{Marcos2018_equivariance}
Diego Marcos, Benjamin Kellenberger, Sylvain Lobry, and Devis Tuia,
\newblock ``{Scale Equivariance in CNNs with Vector Fields},''
\newblock in {\em Proceedings of the International Conference on Machine
  Learning, ICML'2018, Workshop on Towards learning with limited labels:
  Equivariance, Invariance, and Beyond}, Stockholm, Sweden, July 2018.

\bibitem{Yu2015_Dilated}
Fisher Yu and Vladlen Koltun,
\newblock ``{Multi-Scale Context Aggregation by Dilated Convolutions},''
\newblock in {\em Proceedings of the International Conference on Learning
  Representations, ICLR'2016}, Yoshua Bengio and Yann LeCun, Eds., San Juan,
  Puerto Rico, 2016.

\bibitem{Szegedy2015_GoogleNet}
Christian Szegedy, Wei Liu, Yangqing Jia, Pierre Sermanet, Scott Reed, Dragomir
  Anguelov, Dumitru Erhan, Vincent Vanhoucke, and Andrew Rabinovich,
\newblock ``{Going Deeper with Convolutions},''
\newblock in {\em Proceedings of the IEEE Conference on Computer Vision and
  Pattern Recognition, CVPR'2015}, Boston, MA, USA, June 2015, pp. 1--9.

\bibitem{Sundararajan2017_IR}
Mukund Sundararajan, Ankur Taly, and Qiqi Yan,
\newblock ``{Axiomatic Attribution for Deep Networks},''
\newblock in {\em Proceedings of the 34th International Conference on Machine
  Learning, ICML'2017}, Sydney, NSW, Australia, Aug. 2017, vol.~70, pp.
  3319--3328.

\end{thebibliography}

\end{document}